\documentclass[
]{ceurart}

\usepackage{listings}
\begin{document}

\copyrightyear{2023}
\copyrightclause{Copyright for this paper by its authors.
  Use permitted under Creative Commons License Attribution 4.0
  International (CC BY 4.0).}

\conference{CLEF 2023: Conference and Labs of the Evaluation Forum, September 18–21, 2023, Thessaloniki, Greece}

\title{Gpachov at CheckThat! 2023: A Diverse Multi-Approach Ensemble for Subjectivity Detection in News Articles}

\title[mode=sub]{Notebook for the CheckThat Lab at CLEF 2023}

\author[1]{Georgi Pachov}[%
email=georgi.patchov@gmail.com,
]
\cormark[1]
\address[1]{Sofia University “St. Kliment Ohridski”, Bulgaria}
\address[2]{Mohamed bin Zayed University of Artificial Intelligence, UAE}

\author[1]{Dimitar Dimitrov}[%
email=mitko.bg.ss@gmail.com,
]

\author[1]{Ivan Koychev}[%
email=koychev@fmi.uni-sofia.bg,
]

\author[2]{Preslav Nakov}[%
email=preslav.nakov@mbzuai.ac.ae,
]

\cortext[1]{Corresponding author.}

\begin{abstract}
The wide-spread use of social networks has given rise to subjective, misleading, and even false information on the Internet. Thus, subjectivity detection can play an important role in ensuring the objectiveness and the quality of a piece of information. This paper presents the solution built by the Gpachov team for the CLEF-2023 CheckThat! lab Task~2 on subjectivity detection.
Three different research directions are explored. The first one is based on fine-tuning a sentence embeddings encoder model and dimensionality reduction. The second one explores a sample-efficient few-shot learning model. The third one evaluates fine-tuning a multilingual transformer on an altered dataset, using data from multiple languages.
Finally, the three approaches are combined in a simple majority voting ensemble, resulting in 0.77 macro F1 on the test set and achieving 2nd place on the English subtask. 

\end{abstract}

\begin{keywords}
Subjectivity detection \sep
Sentence Embeddings \sep
Few-shot learning \sep
Transformer \sep
Ensemble \sep
Natural Language Processing \sep
Deep Learning
\end{keywords}

\maketitle

\section{Introduction}

Subjectivity is a feature of language and a form of bias, in which whenever a person is sharing information, it comes out skewed by the speaker’s own personal preferences, beliefs and views. In today’s interconnected world, where opinions and biases travel fast and far, subjectivity detection can be a very important piece in order to ensure information reporting is done in a clear, objective and unbiased fashion.

Specifically, subjectivity in news and media articles can be nuanced, subtle and difficult to identify. Detection of subjectivity in such texts can play an important role in identifying potentially misleading or malicious texts and in detecting fake news online.

In Task 2 of CheckThat! Lab at CLEF 2023 \cite{clef-checkthat:2023, clef-checkthat:2023:task2}, systems are required to distinguish whether a sentence from a news article expresses the subjective view of the author or presents an objective view on the covered topic instead. This is a binary classification task in which systems have to identify whether a text (a sentence or a paragraph) is subjective or objective.

This paper explores the effects of fine-tuning a large pre-trained language model on the subjectivity task. Additionally, we have examined the angles of few-shot-learning and fine-tuning a sentence embedding model. Finally, an ensemble method is proposed to unify all three into one solution.

\section{Related Work}
While sentiment analysis can be regarded as a classic NLP task with lots of research already available on the subject,  subjectivity classification is deemed to be a slightly less popular research topic. \cite{CHATURVEDI201865} looks at the subjectivity detection task as a way to improve sentiment analysis classifiers by excluding neutral (objective) sentences. They offer a broad survey on published subjectivity detection methods, categorizing them into syntactic (keyword-spotting, lexical affinity, statistical methods), semantic (parse trees, convolutional neural networks, extreme learning machines) and multi-modal (BiLSTM, multiple-kernel learning). 

In \cite{satapathy2022polarity}, authors explore multi-task learning with hard parameter sharing via Neural Tensor Network. They demonstrate that using a single network with shared layers while learning on two semantically related datasets can improve performance on both datasets.

In \cite{inbook}, authors compare Word2Vec and BERT embedding models in the context of subjectivity detection. With additional classification models to process the embedding outputs, authors demonstrate the superiority of BERT embeddings in high-resource settings, while showing Word2Vec embeddings can be more efficient in low-resource settings. In the current challenge, pairing an embedding encoder with various classification models is also explored. 

In \cite{karl2022transformers}, authors compare the performance of pure transformer models against a variety of more specialized methods for short text classification. Their results show superior performance of the transformer models and part of the research performed in this paper is influenced by their findings.

\section{Data and Baseline Solution}
For the subjectivity detection task, datasets in 6 different languages were provided - Arabic, Dutch, English, German, Italian and Turkish. A total of 7 datasets were available - one for each language, and one for the multilingual version of the task. 
The English dataset contained a total of 1019 examples. 800 of the provided examples were labeled as training, the other 219 as validation. A baseline solution\footnote{\url{https://gitlab.com/checkthat_lab/clef2023-checkthat-lab/-/tree/main/task2/baseline}} is provided by competition organizers, which consists of a sentence encoder model, producing sentence embeddings, which are then classified with Logistic Regression.

An interesting insight was that most of the sequences were 
relatively short. For the English dataset, the average number of 
words in a sequence was 23, while 90\% of sequences consisted of 40 words or less.

The English training set is imbalanced, with 64\% of samples labeled objective and 36\% - subjective. This imbalance is not present in the validation set.

\section{Experiments and Evaluation}
Three research directions were explored, each of them resulting in a separate solution. The final program used for submission is a simple majority voting ensemble of the three solutions. 
The first research direction explores what is achievable using sentence embeddings. The second one looks at a few-shot-learning model and dual-stage fine-tuning. The third is based on fine-tuning a pre-trained transformer model, also utilizing training data from the other languages available for the task.

All evaluations of experiments are done on the English validation set, provided by the organizers. All research directions will be described in more detail in the next subsections.

\subsection{Sentence Embeddings}
Multiple experiments with sentence embeddings were conducted. All of them were based on using a pre-trained sentence embedding encoder model \cite{reimers2019sentencebert} . The following ideas were explored:

\begin{itemize}
\item Using more powerful classifiers on top of sentence embeddings output
\item Using dimensionality reduction
\item Fine-tuning the sentence embeddings encoder in the context of subjectivity detection
\end{itemize}

Initially, the baseline solution provided by organizers used sentence embeddings and a simple Logistic Regression on top to produce classification outputs. More complex classifiers were tested. Multiple different classifiers yielded improvements, measured on the validation set.  The most performant was LogisticRegression (from sklearn), using ElasticNet penalty,  balanced class weights, ‘saga’ solver and 0.5 as regularization constant.

A potential place for improvement was related to sentence embeddings dimensionality. Due to only having 800 training examples, embeddings of dimensionality 384 could prove challenging for a classifier. Using dimensionality reduction, information from embedding vectors can be further compressed in a way that could make it easier for classifiers to find a proper decision boundary. 

Best performance was achieved using PCA with 110 remaining components (out of 384), which explained ~92.5\% of total variance. Experiments with different classifiers and dimensionality reduction are summarized in the first column of Table \ref{tab:table_2}.


While dimensionality reduction can help classifiers, the sentence embeddings themselves were generally created for a very broad category of NLP tasks. Fine-tuning the embeddings encoder for subjectivity detection proved to be helpful for all of the tested classifiers.

Embeddings were fine-tuned using cosine similarity loss in a contrastive learning manner.  The new similarity label of a pair of sentences consisted of two equally weighted components - their original similarity and their label-based similarity. Label-based similarity is defined as 1 if the two sentences are from the same class and 0 otherwise. This can be summarized with the following equation:

\begin{equation} \label{eq:1}
  New\_Similarity\_Label(A, B) = 0.5 * Similarity(A,B)  + 0.5 * (class(A)==class(B))
\end{equation}

To generate training samples, N objective and N subjective sentences were randomly selected. Training pairs were generated - each sentence was paired with every other sentence and a similarity label was generated using Formula~\ref{eq:1}. In total, 2N*(2N-1) training pairs were generated. 

All of the tested classifiers performed better using the fine-tuned embeddings (with N=100) instead of original embeddings. All experiments with classifiers, dimensionality reduction and fine-tuned sentence embeddings are summarized in Table~\ref{tab:table_2}. For both the original and the fine-tuned embeddings, best performance was achieved through PCA and Linear Regression with elastic net penalty.  All macro F1 scores are measured on the (English) validation set.

\begin{table*}
  \caption{Classifiers and dimensionality reduction with original vs fine-tuned embeddings encoder (Macro F1)}
  \label{tab:table_2}
  \begin{tabular}{ccc}
    \toprule
    Classifier$\backslash$Embedding & Original Sentence Embeddings & Fine-Tuned Embeddings (N=100)\\
    \midrule
    Baseline (SBERT + LR) & 0.74 & 0.76 \\
        SVM & 0.76 & 0.78 \\
        ElasticNet & 0.77 & 0.8 \\
        PCA + ElasticNet & \textbf{0.78} & \textbf{0.81} \\
  \bottomrule
\end{tabular}
\end{table*}

\subsection{Few-Shot Learning}
The second research direction explored what can be achieved with few-shot learning. Experiments are based on the SetFit model from HuggingFace \cite{tunstall2022efficient}. While conceptually similar to the idea of fine-tuning sentence embeddings, the SetFit model has numerous advantages, including faster training, better sample efficiency and a dual-stage fine-tuning mechanism. In the first stage, the classification head is frozen and embeddings are fine-tuned. Vice versa in the second stage.

Experiments and results for this approach are outlined in Table~\ref{tab:table_3}. Results are similar to fine-tuning sentence embeddings encoder, but achieved while using a lot less information (samples) from the task-specific dataset, which showed promise of low variance and good generalization capabilities.

\begin{table*}
  \caption{Few-shot learning experiments with SetFit}
  \label{tab:table_3}
  \begin{tabular}{ccc}
    \toprule
    Fine-tuning regime & Number of samples & Macro F1 \\
    \midrule
        Single stage & 10 & 0.79 \\
        Single stage & 20 & 0.8 \\ 
        Dual stage & 20 & \textbf{0.81} \\ 
        Dual stage & 64 & 0.8 \\ 
        Dual stage & 100 & 0.79 \\ 
  \bottomrule
\end{tabular}
\end{table*}

\subsection{Transformer Fine-Tuning}
The third research direction was based on the idea of fine-tuning a transformer model to the dataset provided. Lots of transformer models are available in the HuggingFace hub \cite{wolf-etal-2020-transformers}. BERT \cite{devlin2019bert}, RoBERTa \cite{liu2019roberta}  and DeBERTa \cite{he2021deberta} were used throughout the experiments. Limitations in hardware capabilities restricted the experiments to only 'base' and 'large' variants of the models. Xlarge and xxlarge models were not tested.  Initially, experiments were performed using the English dataset only. Later on, additional experiments were performed including data from other languages.

Table~\ref{tab:table_4} summarizes the experiments and results when using the English dataset. Fine-tuning a DeBERTa-v2-large  achieved the best performance on the validation set.

\begin{table*}
  \caption{Transformer experiments with English dataset}
  \label{tab:table_4}
  \begin{tabular}{ccc}
    \toprule
    Transformer Model & Architecture & Macro F1  \\
    \midrule
        BERT & base & 0.77 \\
        BERT & large & 0.78 \\
        RoBERTa & base & 0.81 \\
        DeBERTa-v2 & large & \textbf{0.82} \\
  \bottomrule
\end{tabular}
\end{table*}

Additional experiments were performed using data from other languages. Multiple training datasets were created. The first one consisted of all the available data for all languages. The second one consisted of English, Arabic and Turkish - the languages who had published baseline solutions with higher F1 than English. The third one contained training samples from English and German translated to English using a neural model \cite{tang2020multilingual}. While creating the datasets, English validation samples were always manually excluded as a final step to prevent data leakage.

For all three datasets, multilingual transformer models were used - bert-base-multilingual \cite{devlin2019bert}, mDeBERTa \cite{he2023debertav3} and xlm-roberta-base \cite{conneau-etal-2020-unsupervised}. 

Table~\ref{tab:table_5} summarizes the results. An xlm-roberta-base model was fine-tuned using all available data, achieving a macro F1 score of 0.83, which showed slight improvement over using English-only dataset. The best results were obtained by using English and German translated to English, fitted to xlm-roberta-base model.

\begin{table*}
  \caption{Transformer experiments with multilingual datasets}
  \label{tab:table_5}
  \begin{tabular}{cccc}
    \toprule
    Model & Architecture & Training Set & F1 macro  \\
    \midrule
        bert-multilingual & base & All data & 0.81 \\
        xlm-roberta & large & All data & 0.81 \\
        mdeberta-v3 & base & All data & 0.82 \\ 
        xlm-roberta & base & All data & 0.83 \\ 
        xlm-roberta & base & English, Arabic and Turkish & 0.82 \\ 
        xlm-roberta & base & English, German translated to English & \textbf{0.84} \\ 
  \bottomrule
\end{tabular}
\end{table*}

\subsection{Ensemble}

\begin{figure}
  \centering
  \includegraphics[width=\linewidth]{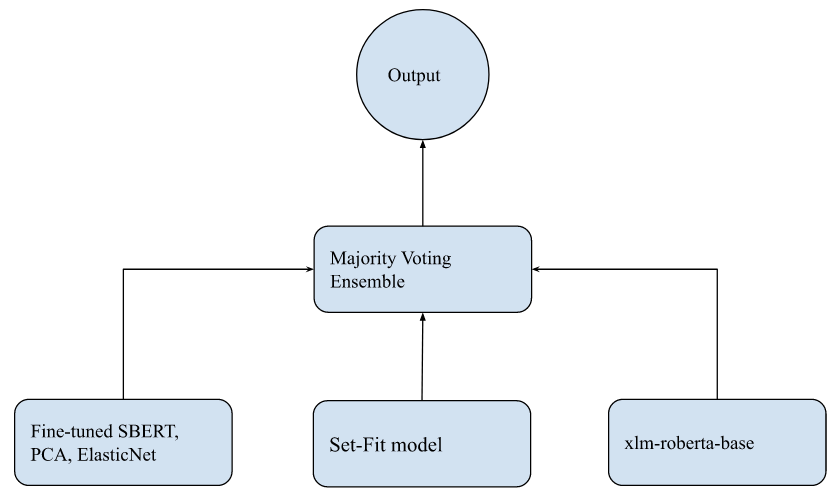}
  \caption{Schema of the ensemble} 
  \label{fig:ensemble}
\end{figure}

Best performing solutions from all three research directions were chosen and formed into an ensemble (Figure~\ref{fig:ensemble}). The final solution was a simple majority voting ensemble from the following solutions:

\begin{itemize}
\item A fine-tuned sentence embeddings encoder, producing improved sentence embeddings (in the context of this task). The encoder output passes through dimensionality reduction (384 dimensions reduced to 110). The reduced embeddings are then classified through LogisticRegression with equally weighted L1 and L2 penalties, ‘saga’ solver, balanced class weights and 0.5 regularization constant.
\item A few-shot learning SetFit model trained using dual-stage fine-tuning procedure with N=20 sentence pairs. 
\item An xlm-roberta-base model, fine-tuned on English and German translated to English.
\end{itemize}

The final submitted ensemble was able to achieve 0.85 macro F1 on English validation set.

\section{Conclusions}

The final solution achieved macro F1 of 0.85 on validation set and 0.77 on test set. This indicates generalization issues, likely stemming from overfitting the solution to the validation set. 

Each of the three separate methods was analyzed on the test set, to identify possible causes for the significantly lower performance on the test set. All components of the solution perform worse on the test set than on the validation set, indicating a more difficult test set. The first method however incurs a bigger loss of performance than the others. This is likely caused by an issue in the fine-tuning procedure, resulting in the encoder model producing highly specialized embeddings, which lose a bigger part of their pretrained semantics than optimal and thus generalize poorly. Less examples used for contrastive learning, smaller learning rate, or using a holdout set to check for generalization issues could have mitigated this problem.

Results also indicate that the best performer is an xlm-roberta-base model trained on English and translated German. All test and validation F1 scores are summarized in Table~\ref{tab:table_6}.

\begin{table*}
  \caption{Test scores of individual solutions and ensemble}
  \label{tab:table_6}
  \begin{tabular}{ccc}
    \toprule
    Solution & Val F1 score & Test F1 score  \\
    \midrule
        Fine-tuned S-BERT, PCA, ElasticNet & 0.81 & 0.7 \\ 
        SetFit model & 0.81 & 0.76 \\ 
        xlm-roberta-base & 0.84 & \textbf{0.79} \\ 
        Majority voting ensemble & \textbf{0.85} & 0.77 \\ 
  \bottomrule
\end{tabular}
\end{table*}

\begin{table*}
  \caption{Per-class precision, recall and f1-scores of the final solution}
  \label{tab:table_7}
  \begin{tabular}{cccc}
    \toprule
    Class & Precision & Recall & F1   \\
    \midrule
        subjective & 0.83 & 0.71 & 0.77 \\
        objective & 0.73 & 0.84 & 0.78 \\
  \bottomrule
\end{tabular}
\end{table*}

Table~\ref{tab:table_7} indicates that the solution is biased to expect more frequent ‘objective’ examples, hence the higher recall but lowered precision. This is likely stemming from the initial class imbalance in the English training set, where 64\% of examples are labeled objective.  While this imbalance doesn’t seem to have a major impact on final F1 scores, a more balanced solution could have been achieved by using any of the well-known class-imbalance techniques, e.g. sample weighing or choosing a different threshold.

Overall, xlm-roberta model trained on English and translated German seem to be the best performer. The few-shot-learning approach yielded decent test results. The fine-tuned SBERT encoder method seems to not be general enough and is reducing the performance of the full solution.  Submitting predictions from only the transformer model would have resulted in 0.79 macro F1, which could have won the English subtask challenge.

\section{Future Work}
As indicated by Table {\ref{tab:table_6}}, the transformer-based solution proved to be the most effective and robust out of the ones used in the ensemble. Further experiments can be conducted with newer and more promising versions of existing transformer models. For example, DeBERTa-V3 \cite{he2023debertav3} is reported to improve performance of the original DeBERTa model with 1.37\% on the GLUE benchmark. This can prove relevant to the task of subjectivity classification as well. 

Additionally, due to resource constraints, transformer models with 'xlarge' and 'xxlarge' architectures were not used for this research. For the same reason, very little hyperparameter tuning was performed. Using a larger model and exploring bigger hyperparameter space can potentially improve the results.

Finally, the transformed-based approach used for subtask 2A (English) can also be attempted and used for the other 5 languages in the task - Arabic, Dutch, German, Italian and Turkish.

\begin{acknowledgments}
This research is partially funded by Project UNITe BG05M2OP001-1.001-0004 funded by the OP "Science and Education for Smart Growth", co-funded by the EU through the ESI Funds, and partially financed by the European Union-NextGenerationEU, through the National Recovery and Resilience Plan of the Republic of Bulgaria, project SUMMIT, No BG-RRP-2.004-0008.
\end{acknowledgments}

\bibliography{sample-ceur}

\appendix

\end{document}